\documentclass[journal, letterpaper]{IEEEtran}

\usepackage{graphicx}

\usepackage{xcolor}
\usepackage{hyperref}

\hypersetup{
    colorlinks=true,   
    linkcolor=blue,    
    citecolor=blue,    
    filecolor=blue,    
    urlcolor=blue      
}

\usepackage{url}

\usepackage{amsmath}

\usepackage{textgreek}	
\usepackage{listings}
\usepackage{csvsimple}
\usepackage{longtable}
\usepackage{url}

\begin{document}

	\title{Weld n'Cut: Automated fabrication of inflatable fabric actuators }

\author{Arman Goshtasbi$^{1\dagger}$, Burcu Seyidoğlu$^{1\dagger}$,  Saravana Prashanth Murali Babu$^{1}$, Aida Parvaresh$^{1}$,\\  Cao Danh Do$^{1}$, Ahmad Rafsanjani$^{1*}$

\thanks{$^{1}$SDU Soft Robotics, Biorobotics Section, The Maersk Mc-Kinney Moller Institute, University of Southern Denmark (SDU), 5230 Odense M, Denmark
        {\tt\small (email: \{argo, bseyi, spmb, aidap, cdd, ahra\}@sdu.dk)}\newline$^{*}$Author to whom correspondence should be addressed.
        \newline$^{\dagger}$These authors contributed equally to this work} 
}
	\markboth{}{}
	\maketitle
    
\begin{abstract}
Lightweight, durable textile-based inflatable soft actuators are widely used in soft robotics, particularly for wearable robots in rehabilitation and in enhancing human performance in demanding jobs. Fabricating these actuators typically involves multiple steps: heat-sealable fabrics are fused with a heat press, and non-stick masking layers define internal chambers. These layers must be carefully removed post-fabrication, often making the process labor-intensive and prone to errors. To address these challenges and improve the accuracy and performance of inflatable actuators, we introduce the Weld n'Cut platform—an open-source, automated manufacturing process that combines ultrasonic welding for fusing textile layers with an oscillating knife for precise cuts, enabling the creation of complex inflatable structures. We demonstrate the machine’s performance across various materials and designs with arbitrarily complex geometries.
\end{abstract}


\section{Introduction}

Soft inflatable actuators have transformed the field of robotics by offering flexibility and adaptability, enabling smooth and natural movements through inflation and deflation~\cite{airwelding,inflatable_review}. These characteristics make the soft inflatable robots well-suited for applications where safety is critical, such as medical devices~\cite{science_ranzani,cianchetti2018biomedical}, wearable technology~\cite{nguyen2020design,Thalmanwearable}, and delicate object manipulation~\cite{banana}. The ability of soft inflatables to exhibit diverse deformation patterns enhances their versatility, allowing them to perform complex tasks across diverse environments~\cite{growing,airwelding,pipe}.
Textiles are one of the most widely used materials in fabricating soft inflatable actuators~\cite{textilesoftrobot}. Due to their flexibility, strength, and lightweight properties, textiles allow for precise control over how the actuators move when inflated~\cite{walshtextile,Kneeexo}. By carefully manipulating textile properties such as its weave~\cite{sanchez20233dknitting}, stretchability~\cite{nguyen2020design}, and pattern~\cite{sciencetextile,kirigami_design}, designers can control the direction and degree of deformation, dictating how the actuator bends, twists, or contracts~\cite{sanchez20233dknitting,nguyen2020design,sciencetextile}. Among these properties, the pattern and stretchability of the textile are especially critical in soft robotics, as they offer a greater range of motion and expanded design possibilities for creating more dynamic, functional actuators.
Despite the importance of textile-based soft inflatables, current fabrication methods are still predominantly manual, which presents several challenges in terms of precision and scalability. One commonly used approach involves using a laser cutter to cut the textile and later fuse the layers with a heat press while using masking layers for creating air pouches~\cite{kirigami_design}. In this method, the laser cutter shapes the textile layers, while the masking layer protects specific areas from being bonded during the heat-pressing process. While this method can produce precise bonds, it is labor-intensive and time-consuming, particularly for complex patterns that require meticulous masking. Removing the masking layer after sealing can be problematic, often leading to imperfections that reduce overall efficiency and precision. Additionally, there is a risk that the intricate masking layers may prevent bonding at critical points, leading to leaks and malfunctioning samples.

\begin{figure}[t]
\centering
\includegraphics[width=\columnwidth]{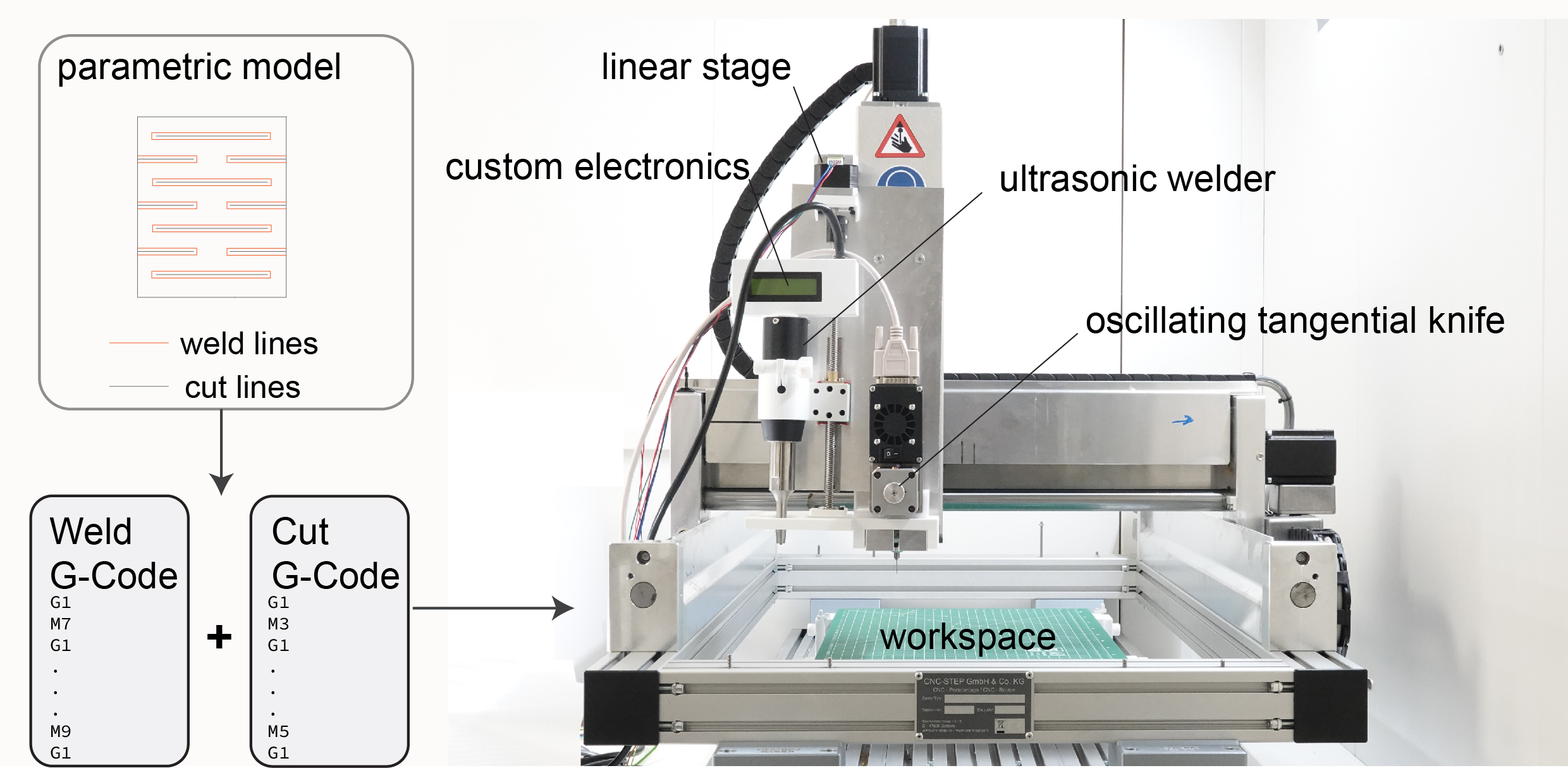}
\caption{The workflow of the \textit{Weld n'Cut} platform.}
\label{Fig1}
\end{figure}

Several studies have proposed alternative methods for fusing textile layers without the need for masking. Amiri Moghadam \textit{et al.} used a laser cutting machine to fuse 37 $\mu$m thin TPU-coated textiles, achieving a variety of motions~\cite{lasercut_1}. Ren \textit{et al.} employed infrared lasers to weld thin polyamide sheets laminated with TPU, creating complex inflatable structures and geometries~\cite{laser_cut2}. While these methods show promise, using laser cutters to fuse textiles introduces limitations on material thickness and the types of materials that can be fused. Furthermore, precise laser tuning is essential, as too much power can cut through the textile, while insufficient power may result in inadequate sealing.
A promising approach for automating the fabrication of inflatable textile actuators involves using computer numerical control (CNC) machines equipped with heat-sealing elements to create complex geometries~\cite{heat_sealing}. Niiyama \textit{et al.}~\cite{sticky_actuator} used a CNC gantry holding a heat
pencil to draw sealing lines on thermoplastic sheets, while Sanchez \textit{et al.}~\cite{airbag} applied a similar technique to fabricate airbags for human-robot interaction. Ou \textit{et al}. \cite{aeromorph} created a large scale robotic sealing platform to create inflatable origami structures and subsequently used a CNC-controlled knife to cut the boundaries. Tools like soldering irons~\cite{soldering} and 3D printing heated extruder heads~\cite{extruder} mounted on CNC gantry have also been explored for fabricating inflatable structures. However, these methods have not yet been extended to include welding and cutting in a single robotic platform. This combination enables the automated fabrication of freeform inflatable kirigami actuators, broadening the functionality of soft robots.
In this work, we introduce the \textit{Weld n'Cut} platform, an open-source automated manufacturing process that utilizes ultrasonic welding to fuse textile layers, combined with an oscillating knife to introduce precise cuts for creating complex inflatable structures. Ultrasonic welding offers significant advantages, including rapid processing, precise control, and the elimination of masking layers, which minimizes fabrication inaccuracies and enhances scalability. The oscillating knife enables the integration of intricate designs, allowing for the creation of dynamic, shape-transforming structures that exhibit more complex deformation behaviors, such as localized expansion, folding, and directional bending.

\begin{figure}[t]
\centering
\includegraphics[width=\columnwidth]{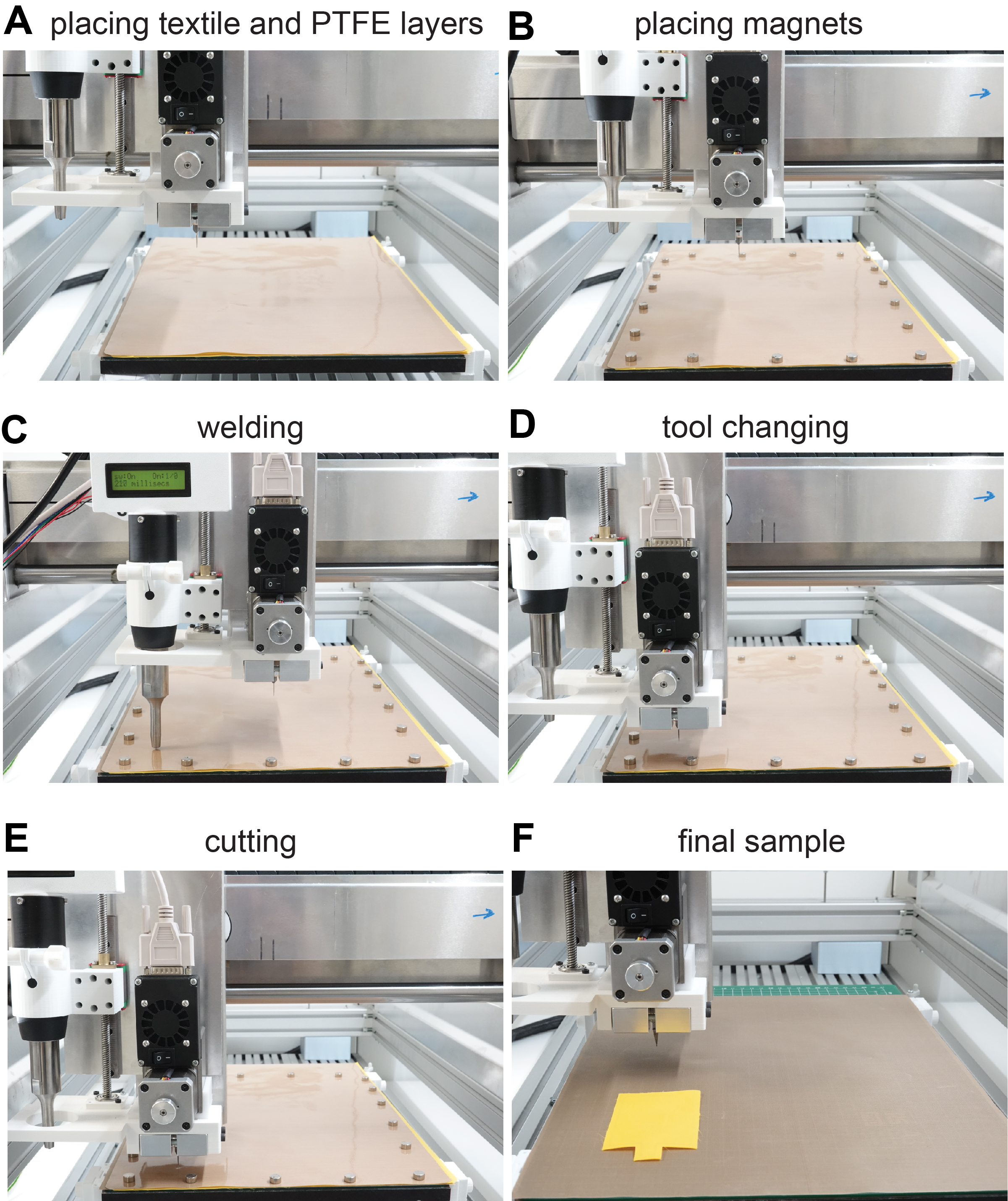}
\caption{fabrication steps. \textbf{A}, Place a PTFE layer on the bottom, followed by two textile layers, and another PTFE layer on top. \textbf{B} Position magnets on top to secure the textiles and PTFE layers in place. \textbf{C} Lower the welder and begin welding the pattern onto the textile. \textbf{D}  Once welding is complete, raise the welder and switch to cutting mode. \textbf{E} Create the cut line with the cutter. \textbf{F} The completed fabricated sample.}
\label{Fig2}
\end{figure}

\section{Materials and Methods}
\label{sec:background}

\subsection{Weld n'Cut Platform}

We developed the Weld n'Cut platform, a custom machine for automating the fabrication of inflatable actuators using lamination techniques and freeform cutting with high precision in a quick and scalable manner. As demonstrated in Fig.~\ref{Fig1}, the platform consists of a 500W ultrasonic plastic spot welder (Baoshishan, China) and an oscillating tangential knife (\href{https://shop.stepcraft.us/shop/11024-oscillating-tangential-knife-otk-3-202808}{Stepcraft, OTK-3}, Germany), both mounted on a Cartesian gantry system (\href{https://www.cnc-step.com/high-z-s-720-cnc-router-720-x-420-x-110-mm-trapezium-screws}{CNC-STEP High-Z S-720}, Germany). We modified the welder’s electronics by designing a circuit that converts it from a manual spot welder to a programmable welder, allowing it to be switched on and off with precise timing intervals (250 ms) without overheating. The two tools are positioned side by side: the knife moves with the z-axis of the CNC gantry, while the welder is attached to a motorized linear stage that allows relative positioning in the vertical direction.
Overall, there are seven channels, all controlled by an open-source machine control software (\href{https://cncdrive.com/UCCNC.html}{UCCNC, CNCDrive}, Hungary). This setup streamlines the fabrication process by enabling control of both welding and cutting tools in a unified system without manually changing tools. (The electronic schematics and CAD files of the holder are on: \href{https://github.com/SDUSoftRobotics/Weldn-cut}{GitHub})

\subsection{Fabrication Workflow}
The workflow begins with the parametric design of welding and cutting profiles. We performed this step using Rhinoceros 7 Grasshopper software. From these designs, we generate the necessary G-codes to control the CNC machine (example file can be found on \href{https://github.com/SDUSoftRobotics/Weldn-cut}{GitHub}). We used a polytetrafluoroethylene (PTFE) sheet (3D Supreme, 0.25mm) on top of the bed to prevent the sticking of the melted materials at high temperatures. The actuator materials are then positioned on top of the PTFE sheet and covered with another PTFE layer to protect the material during welding. The machine bed is composed of a rigid insulating material designed to withstand both the mechanical load from the ultrasonic welder and the heat generated during the welding process. To protect the oscillating knife from damage during cutting, we placed a cutting mat on top of the insulating bed. Fig.~\ref{Fig2}A
For the welding step, the ultrasonic welder is lowered into position using the linear stage, and the programmed welding patterns are executed. Once welding is complete, the welder is raised, and the cutting tool is activated. The cutting process follows the designated patterns, precisely adding the necessary cuts to complete the fabrication (Fig.~\ref{Fig2}B-F).

\subsection{Materials Selection}
The \textit{Weld n'Cut} platform can bond heat-sealable fabrics without the need for stitching or adhesives. These fabrics often have a layer of thermoplastic polymer such as polypropylene (PP), polyethylene (PE), nylon, and thermoplastic polyurethane (TPU) that melts when exposed to heat, allowing two pieces of fabric to fuse together. Unlike the heat-press technique, ultrasonic welding does not require applying high pressure for bonding. It only makes contact with the materials being joined, and bonding occurs under slight pressure. During this process, two fabric layers were joined with their thermoplastic coated side facing each other. The welder generates ultrasonic sound waves, which create enough heat to melt the thermoplastic layer. Once the fabrics cool down, the layers form a durable, waterproof, and airtight bond.
We evaluated a range of fabric types for their welding compatibility, including conductive fabric (Adafruit Velostat with volume Resistivity < 500 ohm-cm), TPU-coated Ripstop (20D, TPU-coated both sides), PU-coated polyester (240 g/sqm), and PU-coated nylon (130 g/sqm). Each material was tried with a standard weld-and-cut geometry, and the bonding performance was tested under pressure up to 50 kPa.  
To optimize the material selection, we also tested TPU-coated nylon in lightweight (170 g/sqm), medium-weight (275 g/sqm), and heavy-weight (450 g/sqm) forms. We ultimately selected TPU-coated nylon due to its strong compatibility with ultrasonic welding, achieving airtight, robust bonds without compromising the material flexibility. Additionally, Nylon's inherent durability and resilience, combined with the benefits of TPU-coating made it an ideal choice for further development. This choice allows the \textit{Weld n'Cut} platform to demonstrate effective welding and cutting for a variety of actuator designs and applications.

\subsection{Actuators Design}

We designed several actuators featuring intricate networks of interconnected air pouches and cut patterns to demonstrate the capabilities of the \textit{Weld n'Cut} platform.
First, we tested simple geometries to illustrate the platform’s ability to create airtight pouches and precisely trim boundaries. We then developed a series of pneumatic network actuators (fabric-based PneuNets) capable of contraction, bending, and twisting movements (see \href{https://github.com/SDUSoftRobotics/Weldn-cut}{GitHub} for G-Codes). For contraction, we used similar fabrics for both layers, whereas for bending and twisting, we fused two layers with different thicknesses and adjusted the geometry of the welding lines to control the movement.
Additionally, we designed an antagonistic bending actuator by fusing three layers to create two separate networks of air pouches that can be actuated independently to achieve bidirectional bending.
Finally, we created a set of complex inflatable kirigami actuators by introducing arrays of sealed cuts. The designs include kirigami actuators with staggered linear cut patterns for uniform contraction, as well as configurations that enable multimodal deformation, allowing for both bidirectional bending and uniform contraction.

\subsection{Actuation}
We actuated the fabricated actuators using a pressure-based flow control system (Flow-EZ, Fluigent, France). We ramped the pressure up to $P=50$ kPa over 20 seconds, then back down to $P=0$ kPa over another 20 seconds, and held at this pressure for another 20 seconds to observe whether each actuator returned to its initial position. Simultaneously, we captured each actuator's response by recording videos using a digital camera (SONY RX-100). Subsequently, we tracked the end-point positions to measure their deformation.

\begin{figure}[t]
\centering
\includegraphics[width=\columnwidth]{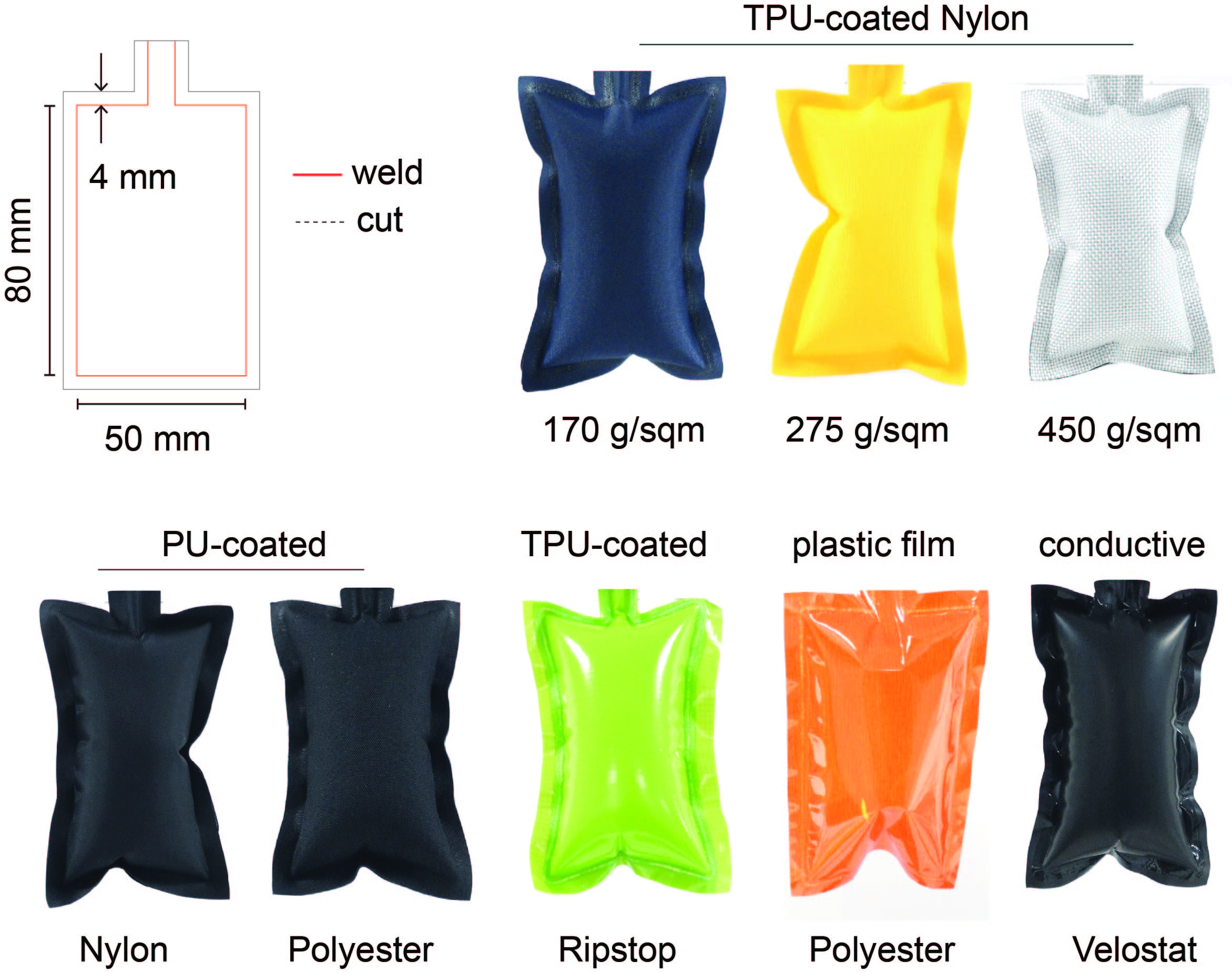}
\caption{Rectangular air pouches fabricated from diverse coated fabrics using the \textit{Weld n'Cut} platform and inflated at $P=50$ kPa. }
\label{Fig3}
\end{figure}

\section{Results and Discussion}
\subsection{Material Exploration}
We fabricated a simple airpouch with a rectangular geometry and an inlet to evaluate the performance of various materials using the \textit{Weld n'Cut} platform (see Fig.~\ref{Fig3}). Each fabric type showed unique bonding characteristics under ultrasonic welding. We then pressurized air pouches up to $P=50$ kPa to ensure their airtightness and bond integrity.
The tested materials include TPU-coated Nylon of different weights, PU-coated Nylon and Polyester, TPU-coated Ripstop, polyester plastic film, and Velostat conductive fabric.
All materials formed bonds with sufficient strength without leakage except the polyester plastic film that burst after inflation. Notably, TPU-coated Ripstop and TPU-coated polyester showed superior performance in comparison to conductive fabric. The conductive fabric is less in weight and it is inherently delicate in comparison to TPU-coated Ripstop and TPU-coated polyester fabric.
The lightweight and mid-weight TPU-coated Nylon provided stable bonds without compromising flexibility while the heavy-weight fabric showed the strongest bonding, as the coating layer is thicker. Based on these findings, TPU-coated Nylon was selected as the primary material for further development. 
To optimize bonding quality, we carefully adjusted the welding speed for each material. Speed is a critical factor, as excessively high speeds can result in weak and discontinuous bonding, while overly slow welding may cause the textile to burn. Based on our observations with various weights of TPU-coated textiles, we found that heavier textiles require lower welding speeds. We achieved optimal bonding with a speed of 200 mm/min for lightweight textiles, 160 mm/min for medium-weight textiles, and 100 mm/min for heavyweight textiles.
For Velostat, which has a low melting temperature, the ideal welding speed was 250 mm/min. For PU-coated textiles, which have a melting temperature similar to TPU, we used the same speed settings according to their weight classification. For PET film, due to its thinness and susceptibility to damage, we added two layers of PTFE for protection and welded at a speed of 120 mm/min.

\subsection{Fabric PneuNets Actuators}
\begin{figure}[t]
\centering
\includegraphics[width=\columnwidth]{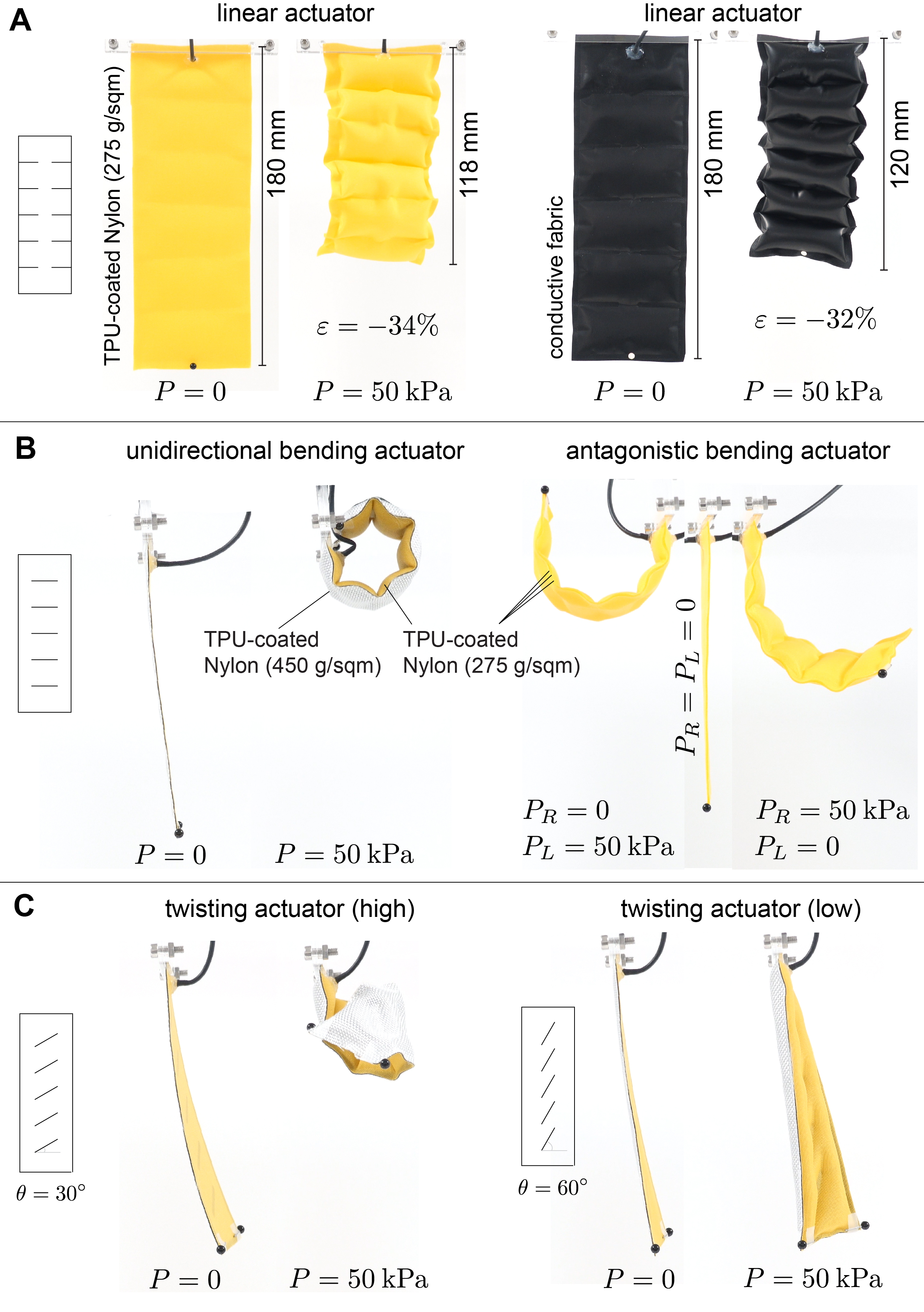}
\caption{Fabric PneuNets actuators. \textbf{A} Linear actuators. \textbf{B} Unidirectional and antagonistic bending actuators. \textbf{C} Twisting actuators with different deformation behavior.}
\label{Fig4}
\end{figure}

We fabricated a series of fabric-based pneumatic networks (PneuNets) actuators to demonstrate the usability of our platform for diverse soft robotics applications.
First, inspired by \cite{nguyen2020design}, we fabricated a linear actuator by welding two layers of mid-weight TPU-coated Nylon (275 g/sqm) featuring a series of weld lines. The outcome is an artificial muscle capable of axial contraction when inflated, which showed a maximum contraction of $\varepsilon=34\%$ at $P=50$ kPa (see Fig.~\ref{Fig4}A, left). We also fabricated the same design using the conductive fabric that can be used to sense contraction, providing proprioceptive feedback. This actuator contracted $\varepsilon=32\%$ at $P=50$ kPa (see Fig.~\ref{Fig4}A, right).
Next, we created a bending actuator by fusing two different weights of TPU-coated Nylon (275 g/sqm and 450 g/sqm) and introducing an array of parallel weld lines. The asymmetry introduced by the contrast in the material stiffness results in bending towards the lighter layer. From a straight line at $P=0$, the actuator turns into a rolled configuration at $P=50$ kPa (see Fig.~\ref{Fig4}B, left). Then, we used the same weld pattern to fuse three layers of the midweight TPU-coated Nylon. This design resulted in an antagonistic actuator with two parallel chambers that can undergo bidirectional bending when each chamber is inflated individually (see Fig.~\ref{Fig4}B, right). 
Finally, we created twisting actuators by fusing two different weights of TPU-coated Nylon (similar to the unidirectional bending actuator) and introducing an array of inclined weld lines with the inclination angles of $\theta=30^\circ$ and $\theta=60^\circ$.
These designs resulted in distinct twisting behaviors as demonstrated in Fig.~\ref{Fig4}C. 

\begin{figure}[t]
\centering
\includegraphics[width=\columnwidth]{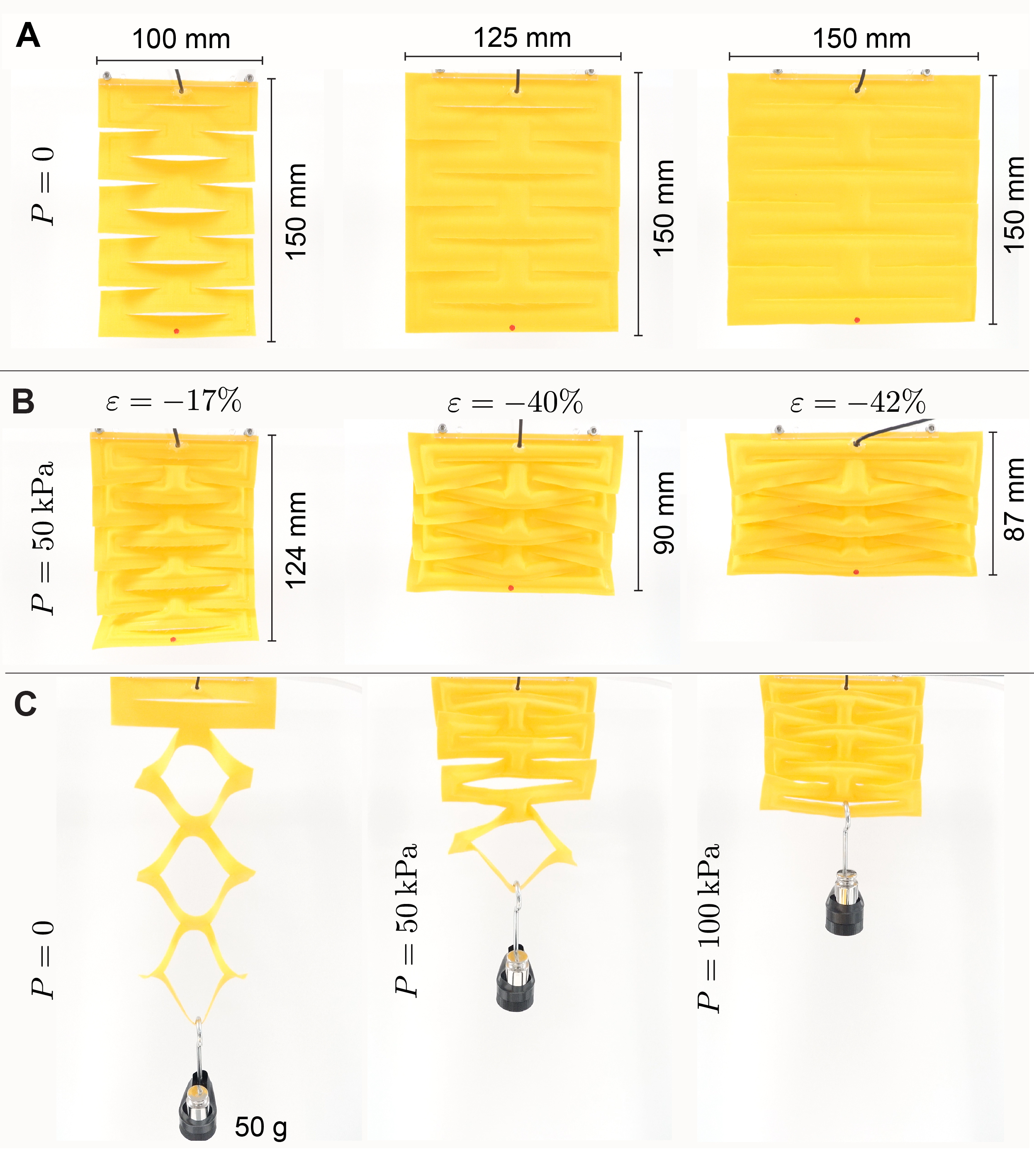}
\caption{Kirigami actuators with a staggered linear cut pattern at different widths with multiple interconnected air channels. ~\textbf{A}. Uninflated, and \textbf{B}. inflated at $P=50$ kPa.
\textbf{C}. using the Kirigami actuator with $w=125$ mm to lift a 50 g weight by applying $P=100$ kPa. }
\label{Fig5}
\end{figure}

\subsection{Kirigami Actuators}
To demonstrate the capability of the \textit{Weld n'Cut} platform in combining precise cutting and robust welding for inflatable structures, we designed a series of complex kirigami actuators comprised of multiple cuts and intricately interconnected channels. Each kirigami actuator featured an array of staggered linear cut patterns.
We considered three different widths ($w=$100, 125, 150 mm) while keeping the height of the actuator constant ($h=$150 mm), which results in internal channels with varying dimensions. The welding was performed in a single continuous channel that ran the entire length of each actuator (see Fig.~\ref{Fig5}A).
We tested these kirigami actuators by gradually increasing the internal pressure up to $P=50$ kPa, monitoring their contraction behavior as an indicator of performance. 
After an initial bulging phase, the actuators axially contract with an overlapping configuration, which is more pronounced for wider cuts. 
This behavior is also reflected in the amount of contraction that they can achieve that was $\varepsilon=-17\%$, $\varepsilon=-40\%$, and $\varepsilon=-42\%$ for $w=100$ mm, $w=125$ mm, and $w=150$ mm, respectively (see Fig.~\ref{Fig5}B). It should be noted that the fabricated length is slightly longer than the design for this actuator due to residual deformations. However, for a fair comparison, we compared their contractions against their designed length. 

We observed that the bonding integrity was maintained even when pressurizing the actuators to 100 kPa, without any leakage or structural failure. To further showcase the capabilities of these kirigami actuators, we conducted a simple weight-lifting test to evaluate their load-carrying potential.
Fig.~\ref{Fig5}C demonstrates that the actuator can successfully lift a 50 g weight when pressurized to $P=100$ kPa. These results demonstrate the platform's ability to reliably produce varying prototypes and structures, highlighting its potential for applications in fabric-based inflatable structures.

\section{Conclusion}

In summary, we presented \textit{Weld n'Cut}, a robotic platform to automate the fabrication of textile-based inflatable actuators for soft robotics applications. By combining ultrasonic welding and precision cutting in a unified system, the developed machine effectively addresses the limitations of manual fabrication, such as misalignment, material deterioration under heat press, and multiple fabrication steps. The platform's capability to process various materials through tailoring welding speed to material characteristics demonstrates its versatility, achieving robust, airtight bonds necessary for reliable actuator performance. Additionally, its ability to create complex actuator designs, such as kirigami-patterned and pneumatic network actuators, highlights the platform’s potential for wide-ranging applications in soft robotics.
For future work, bonding characterization for different materials should be conducted to identify optimized parameters, including switching frequency and applied load. Additionally, the process should be refined to prevent cuts in the PTFE sheets, allowing them to be reused multiple times. Developing an automated system for attaching the connectors would close the loop to create functional inflatable actuators in one shot.

\section{Acknowledgement}
This work was supported by the Independent Research Fund Denmark through Sapere Aude grant 1051-00075B and the Villum Young Investigator grant 37499.

\appendix
The supporting video of this article can be found at \url{https://youtu.be/cfQpgHbBx4o}. Details on the electronics layout, Rhino Grasshopper files for generating G-codes, and SolidWorks files for designing holders are available in the GitHub repository 
\url{https://github.com/SDUSoftRobotics/Weldn-cut}.

\bibliographystyle{IEEEtrans}
\bibliography{reference} 

\end{document}